\documentclass[11pt,a4paper]{article}
\usepackage[hyperref]{acl2020}
\usepackage{times}
\usepackage{latexsym}

\usepackage{helvet}  %Required
\usepackage{courier}  %Required
\usepackage{url}  %Required
\usepackage{graphicx}  %Required
\usepackage{mathtools}
\usepackage{amsmath,amsfonts,amsthm,amssymb}
\usepackage{mathrsfs}
\usepackage{subfigure}
\usepackage{booktabs}
\usepackage[utf8]{inputenc}
\usepackage{multirow}

\aclfinalcopy % Uncomment this line for the final submission
\def\aclpaperid{514} %  Enter the acl Paper ID here

\title{Distinguish Confusing Law Articles for Legal Judgment Prediction}

\author{
  Nuo Xu$^1$, Pinghui Wang$^{2, 1}$\footnotemark[1], Long Chen$^1$,
  Li Pan$^3$, Xiaoyan Wang$^4$, Junzhou Zhao$^1$\footnotemark[1]\\
  $^1$MOE NEKEY Lab, Xi'an Jiaotong University, China \\
  $^2$Shenzhen Research School, Xi'an Jiaotong University, China \\
  $^3$School of Electronic, Information and Electrical Engineering, Shanghai Jiao Tong University \\
  $^4$Information Technology Service Center, The Supreme People's Court, China \\
  \texttt{nxu@sei.xjtu.edu.cn, phwang@mail.xjtu.edu.cn,} \\
  \texttt{chenlongche@stu.edu.cn, panli@sjtu.edu.cn,} \\
  \texttt{wangxiaoyan@court.gov.cn, junzhouzhao@gmail.com} \\
  }

\date{}

\begin{document}

\maketitle

\renewcommand{\thefootnote}{\fnsymbol{footnote}}
\footnotetext[1]{Corresponding authors.}
\renewcommand{\thefootnote}{\arabic{footnote}}

\begin{abstract}
Legal Judgment Prediction (LJP) is the task of automatically predicting a law case's judgment results given a text describing its facts, which has excellent prospects in judicial assistance systems and convenient services for the public.
In practice, confusing charges are frequent, because law cases applicable to similar law articles are easily misjudged.
For addressing this issue, the existing method relies heavily on domain experts, which hinders its application in different law systems.
In this paper, we present an end-to-end model, \textit{LADAN}, to solve the task of LJP.
To distinguish confusing charges, we propose a novel graph neural network to automatically learn subtle differences between confusing law articles and design a novel attention mechanism that fully exploits the learned differences to extract compelling discriminative features from fact descriptions attentively.
Experiments conducted on real-world datasets demonstrate the superiority of our LADAN.
\end{abstract}

\section{Introduction} \label{sec:introduction}
\begin{figure}[t]
\centering
\includegraphics[width=1.0\linewidth]{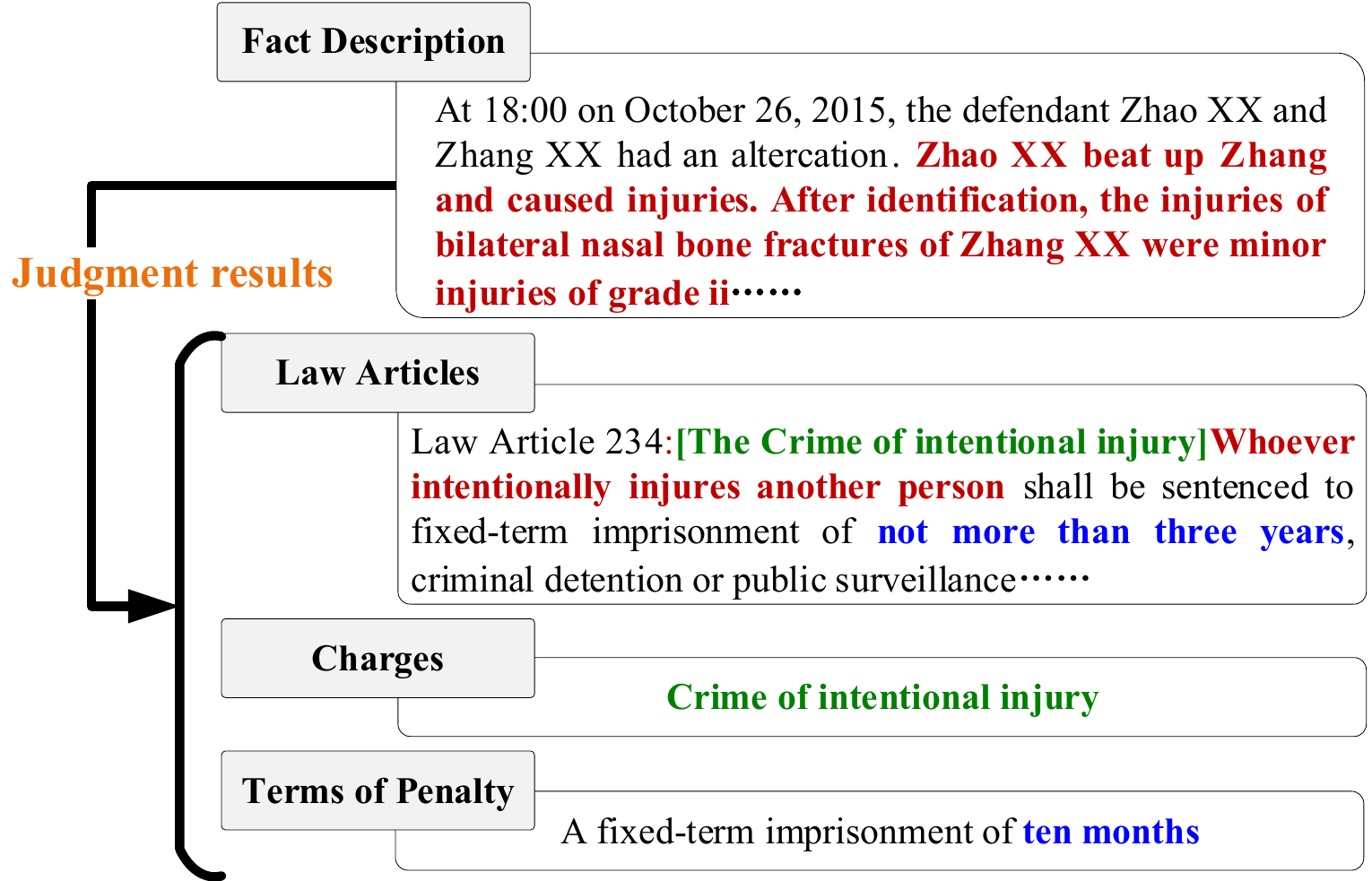}
\caption{An illustration of the LJP. Generally, a judge needs to conduct professional analysis and reasoning on the fact description of the case, and then choose reasonable law articles, charges and the term of penalty to convict the offender.}
\label{fig:0}
\end{figure}

Exploiting artificial intelligence techniques to assist legal judgment has become
popular in recent years.
Legal judgment prediction (\textbf{LJP})
aims to predict a case's judgment results, such as applicable law articles,
charges, and terms of penalty, based on its fact description, as illustrated in
Figure~\ref{fig:0}.
LJP can assist judiciary workers in processing cases and offer legal consultancy
services to the public.
In the literature, LJP is usually formulated as a text classification problem, and
several rule-based methods~\cite{liu2004case,lin2012exploiting} and neural-based
methods~\cite{hu2018few,luo2017learning,zhong2018legal} have been proposed.

\begin{figure}[t]
\centering
\includegraphics[width=1.0\linewidth]{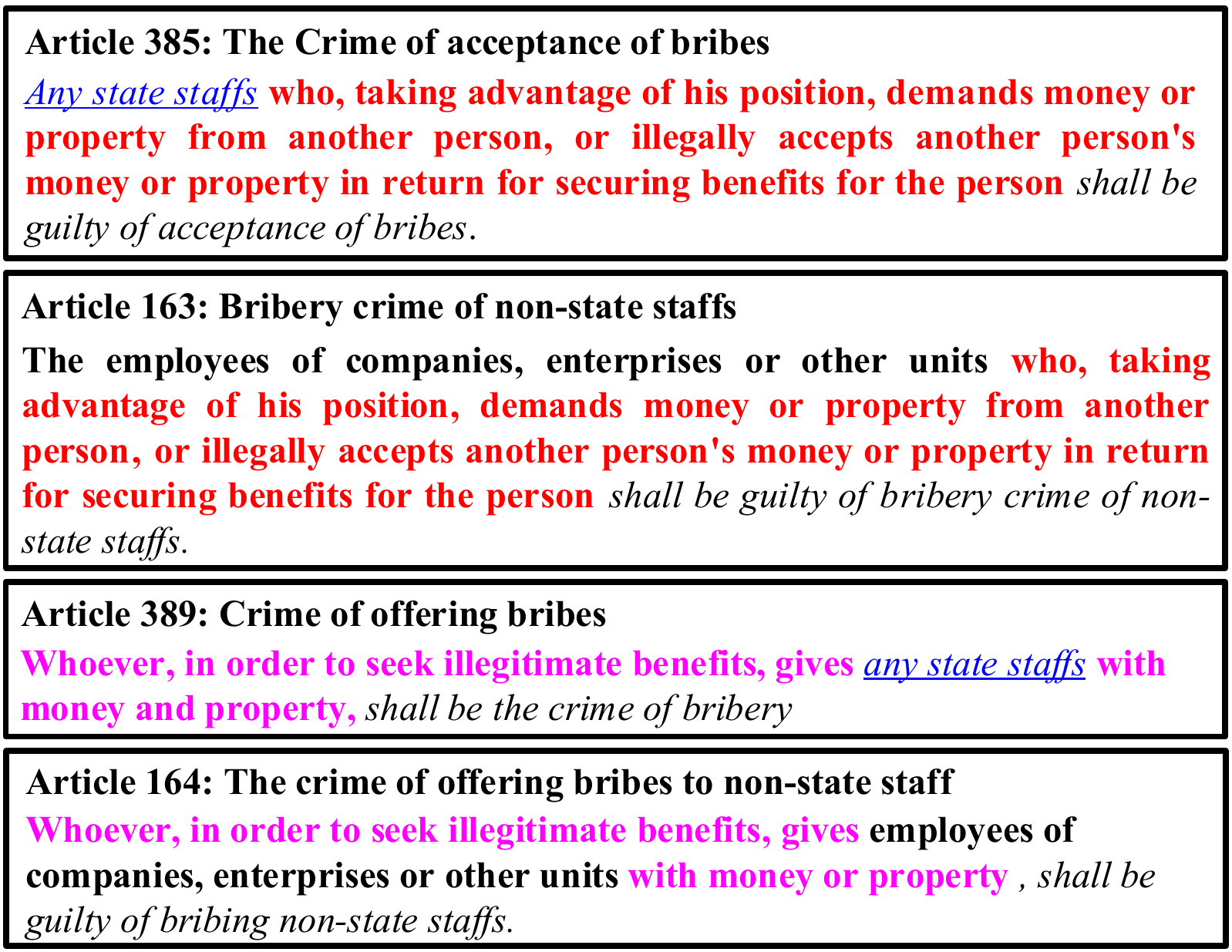}
\caption{Examples of confusing charges.}
\label{fig:law_artices}
\end{figure}

\begin{figure}[t]
\centering
\includegraphics[width=1.0\linewidth]{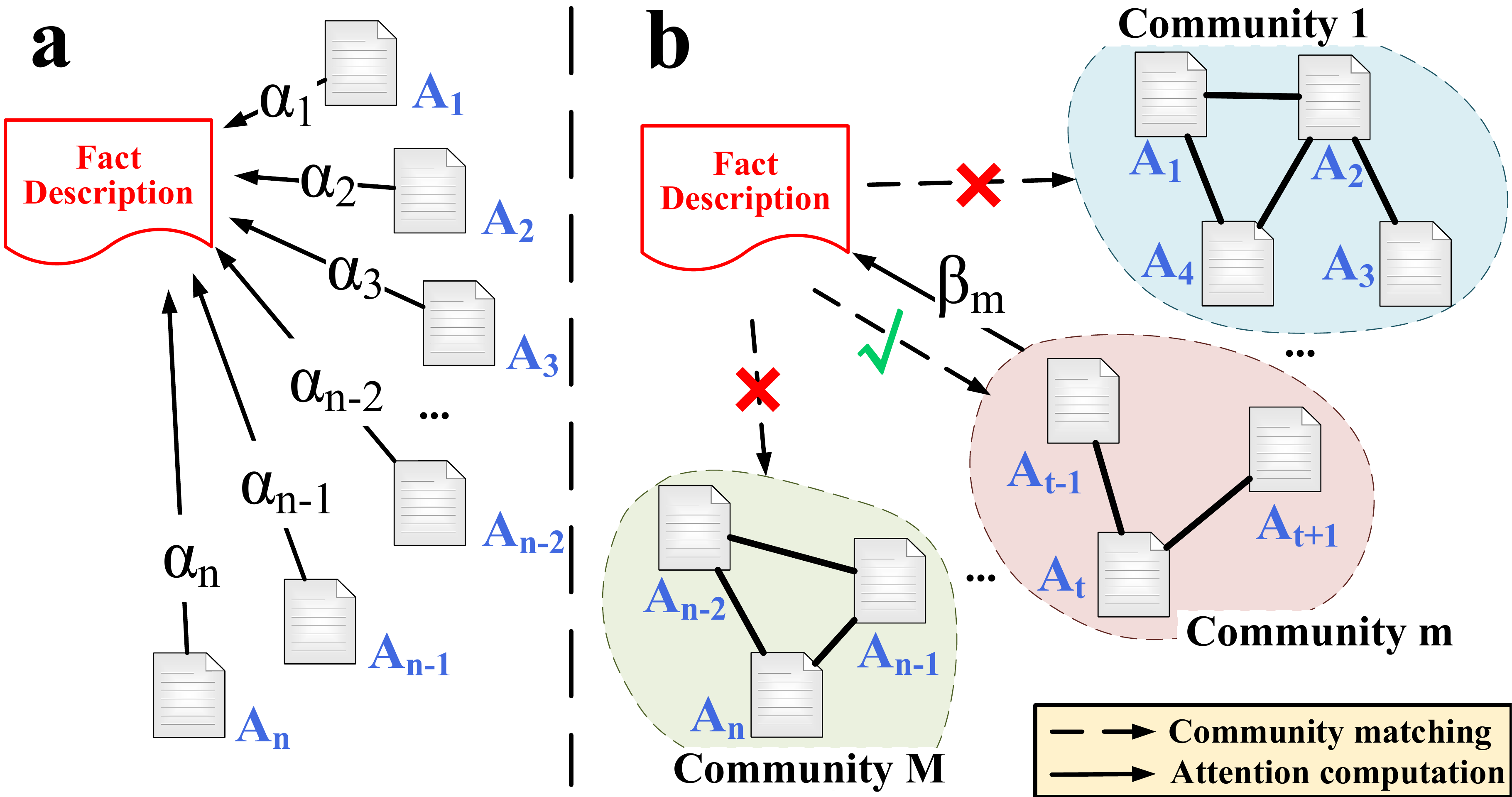}
\caption{\textbf{a.}~The fact-law attention model in ~\cite{luo2017learning}.
  \textbf{b}.~Our framework.
  \textit{Variables $\mathbf{\alpha}$ and $\mathbf{\beta}$ represent the encoded
    vectors learned for attentively extracting features from fact
    descriptions.}}
\label{fig:attention_frameworks}
\end{figure}

The main drawback of existing methods is that they cannot solve the {\em confusing
  charges issue}.
That is, due to the high similarity of several law articles, their corresponding law
cases can be easily misjudged.
For example, in Figure~\ref{fig:law_artices}, both \textit{Article 385} and
\textit{Article 163} describe offenses of accepting bribes, and their subtle
difference is whether the guilty parties are state staffs or not.
The key to solving the confusing charges issue is how to capture essential but rare
features for distinguishing confusing law articles.
Hu et al.~\shortcite{hu2018few} defined ten discriminative attributes to
distinguish confusing charges.
However, their method relies too much on experts to hinder its applications
in a large number of laws.
In practice, we desire a method that can automatically extract textual features
from law articles to assist JLP.
The most relevant existing work to this requirement is~\cite{luo2017learning}, which used an attention mechanism to extract features from fact descriptions with respect to a specific law article.
As shown in Figure~\ref{fig:attention_frameworks}a, for each law article, an
attention vector is computed, which is used to extract features from the fact description of a law case to predict whether the law article is applicable to the case.
Nevertheless, the weakness is that they learn each law article's attention vector
independently, and this may result in that similar attention vectors are learned
for semantically close law articles; hence, it is ineffective in distinguishing
confusing charges.

To solve the confusing charges issue, we propose an end-to-end framework, i.e.,
Law Article Distillation based Attention Network (LADAN).
LADAN uses the difference among similar law articles to attentively extract
features from law cases' fact descriptions, which is more effective in
distinguishing confusing law articles, and improve the performance of LJP.
To obtain the difference among similar law articles, a straightforward way is to
remove duplicated texts between two law articles and only use the
leftover texts for the attention mechanism.
However, we find that this method may generate the same leftover texts for
different law article, and generate misleading information to LJP.
As shown in Fig.~\ref{fig:law_artices}, if we remove the duplicated phrases
and sentences between Article 163 and Article 385 (i.e., the red text in
Fig.~\ref{fig:law_artices}), and between Article 164 and Article 389 (i.e., the pink
text in Fig.~\ref{fig:law_artices}), respectively, then Article 385 and Article
389 will be almost same to each other (i.e., the blue text in
Fig.~\ref{fig:law_artices}).

We design LADAN based on the following observation: it is usually easy to
distinguish dissimilar law articles as sufficient distinctions exist, but challenging to
discriminate similar law articles due to the few useful features.
We first group law articles into different communities, and law articles in the
same community are highly similar to each other.
Then we propose a graph-based representation learning method to automatically
explore the difference among law articles and compute an attention vector for each
community.
For an input law case, we learn both macro- and micro-level features.
Macro-level features are used for predicting which community includes the
applicable law articles.
Micro-level features are attentively extracted by the attention vector of the
selected community for distinguishing confusing law articles within the same
community.
Our main contributions are summarized as follows:

(1) We develop an end-to-end framework, i.e., LADAN, to solve the LJP task.
It addresses the confusing charges issue by mining similarities between fact
descriptions and law articles as well as the distinctions between confusing law articles.

(2) We propose a novel graph distillation operator
(GDO) to extract discriminative features for effectively distinguishing confusing law articles.

(3) We conduct extensive experiments on real-world datasets.
The results show that our model outperforms all
state-of-the-art methods.

\section{Related Work} \label{sec:related}
Our work solves the problem of the confusing charge in the LJP task by referring to the calculation principle of graph neural network (GNN).
Therefore, in this section, we will introduce related works from these two aspects.
\subsection{Legal Judgment Prediction}
Existing approaches for legal judgment prediction (LJP) are mainly divided into three categories.
In early times, works usually focus on analyzing existing legal cases in specific scenarios with mathematical and statistical algorithms~\cite{kort1957predicting,nagel1963,keown1980mathematical,lauderdale2012supreme}. However, these methods are limited to small datasets with few labels.
Later, a number of machine learning-based methods \cite{lin2012exploiting,liu2004case,sulea2017exploring} were developed to solve the problem of LJP, which almost combine some manually designed features with a linear classifier to improve the performance of case classification.
The shortcoming is that these methods rely heavily on manual features, which suffer from the generalization problem.

In recent years, researchers tend to exploit neural networks to solve LJP tasks.
Luo et al.~\shortcite{luo2017learning} propose a hierarchical attentional network to capture the relation between fact description and relevant law articles to improve the charge prediction.
% More recent works research the dependencies of the various subtasks of the JLP problem.
Zhong et al.~\shortcite{zhong2018legal} model the explicit dependencies among subtasks with scalable directed acyclic graph forms and propose a topological multi-task learning framework for effectively solving these subtasks together.
Yang et al.~\shortcite{yang2019legal} further refine this framework by adding backward dependencies between the prediction results of subtasks.
To the best of our knowledge, Hu et al.~\shortcite{hu2018few} are the first to study the problem of discriminating confusing charges for automatically predicting applicable charges.
They manually define 10 discriminative attributes and propose to enhance the representation of the case fact description by learning these attributes.
This method relies too much on experts
and cannot be easily extended to different law systems.
To solve this issue, we propose a novel attention framework that automatically extracts differences between similar law articles to enhance the representation of fact description.

\subsection{Graph Neural Network}
Due to its excellent performance in graph structure data, GNN has attracted significant attention~\cite{kipf2016semi, hamilton2017inductive, DSE2019exploring}.
In general, existing GNNs focus on proposing different aggregation schemes to fuse features from the neighborhood of each node in the graph for extracting richer and more comprehensive information:
Kipf et al.~\shortcite{kipf2016semi} propose graph convolution networks which use mean pooling to pool neighborhood information;
GraphSAGE~\cite{hamilton2017inductive} concatenates the node's features and applies mean/max/LSTM operators to pool neighborhood information for inductively learning node embeddings;
MR-GNN~\cite{xu2019mr} aggregates the multi-resolution features of each node to exploit node information, subgraph information, and global information together;
Besides, Message Passing Neural Networks~\cite{gilmer2017neural} further consider edge information when doing the aggregation.
However, the aggregation schemes lead to the over-smoothing issue of graph neural networks~\cite{li2018deeper}, i.e., the aggregated node representations would become indistinguishable, which is entirely contrary to our goal of extracting distinguishable information.
So in this paper, we propose our distillation operation, based on a distillation strategy instead of aggregation schemes, to extract the distinguishable features between similar law articles.

\section{Problem Formulation} \label{sec:problem}

\begin{figure*}[t]
\centering
\includegraphics[width=0.98\linewidth]{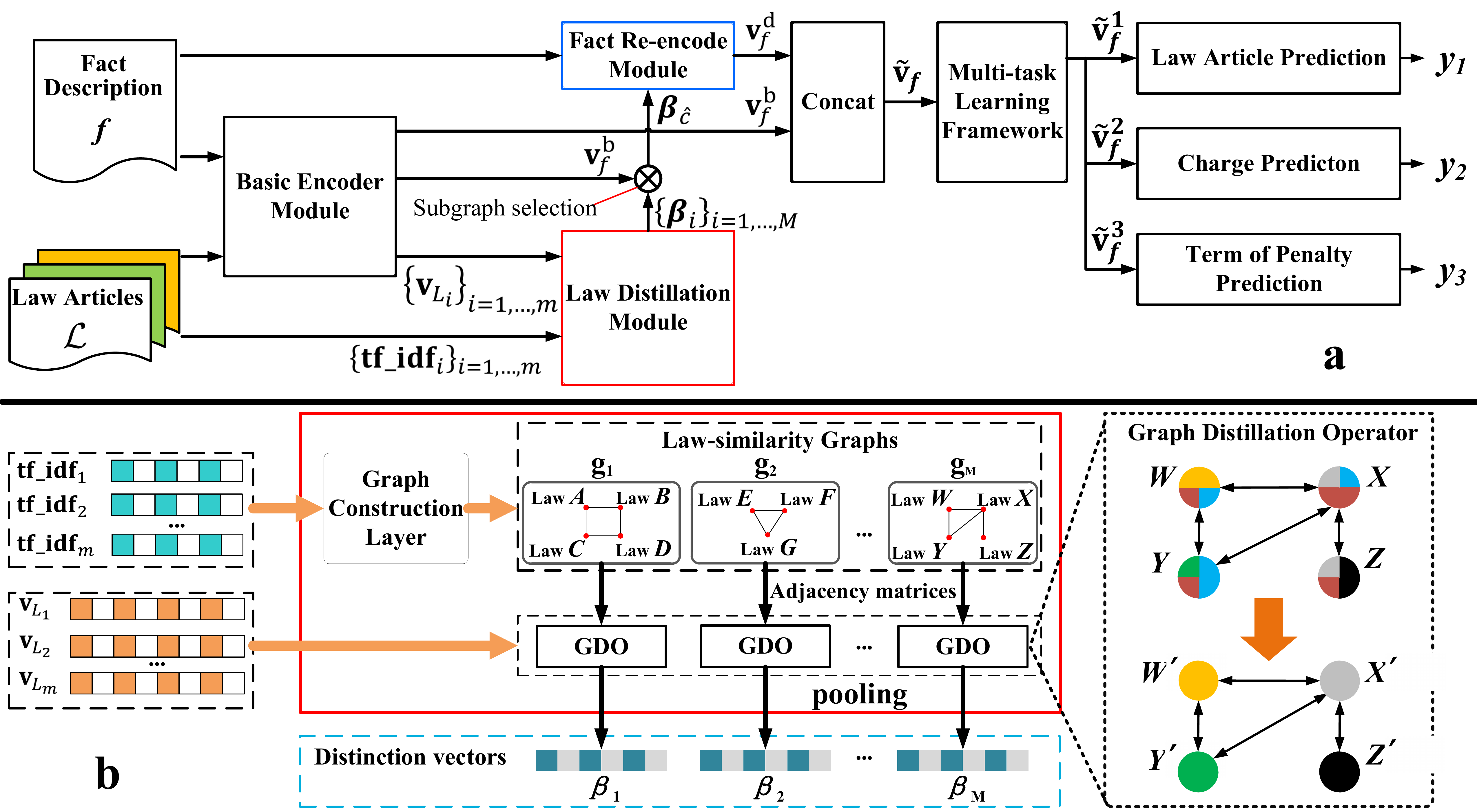}
\caption{\textbf{a.} Overview of our framework \textit{LADAN}: it takes the fact descriptions of cases and the text definitions of law articles as inputs, then extracts the basic representation $\mathbf{v}_f^{\text{b}}$ and distinguishing representation $\mathbf{v}_f^{\text{d}}$ of the fact descriptions through the basic encoder and the re-encoder, and finally combines this two representations for the downstream prediction tasks;
    \textbf{b.} Law Distillation Module: this module communizes law articles and distills the distinguishable features of each community for attention calculation of the re-encoder.}
\label{fig:3}
\end{figure*}

In this section, we introduce some notations and terminologies, and then formulate the LJP task.

\paragraph{Law Cases.}
Each law case consists of a \textit{fact description} and several \textit{judgment results} (cf.~Figure~\ref{fig:0}).
The fact description is represented as a text document, denoted by $f$.
The judgment results may include \textit{applicable law articles}, \textit{charges}, \textit{terms of penalty}, etc.
Assume there are $t$ kinds of judgment results, and the $i$-th judgment result is
represented as a categorical variable $y_i$ which takes value from set $Y_i$.
% Let $Y_i$ be the set of possible judgment results (i.e., labels) of aspect $i$.
Then, a law case can be represented by a tuple $(f, y_1, \ldots, y_t)$.

\paragraph{Law Articles.}
Law cases are often analyzed and adjudicated according to a legislature's \textit{statutory law} (also known as, \textit{written law}).
Formally, we denote the statutory law as a set of \textit{law articles} $\mathcal{L}\triangleq\{L_1,\ldots,L_m\}$ where $m$ is the number of law articles.
Similar to the fact description of cases, we also represent each law article $L_i$ as a document.

\paragraph{Legal Judgment Prediction.}
In this paper, we consider three kinds of judgment results: \textit{applicable law articles}, \textit{charges}, and \textit{terms of penalty}.
Given a training dataset $D\triangleq \{(f, y_1, y_2, y_3)_z\}_{z=1}^{q}$ of size $q$,
we aim to train a model $\text{F}(\cdot)$ that can predict the judgment results for any test law case with a fact description $f_{test}$, i.e., $\text{F}(f_{test}, \mathcal{L}) = (\hat{y}_1, \hat{y}_2, \hat{y}_3)$, where $\hat y_i\in Y_i$, $i=1,2,3$.
Following~\cite{zhong2018legal,yang2019legal}, we assume each case has only one applicable law article.

\section{Our Method} \label{sec:method}

\subsection{Overview}\label{sec:overview}
In our framework LADAN (cf.~Fig.~\ref{fig:3}a), the fact description of a case is
represented by two parts: a \emph{basic representation}, denoted by $\mathbf{v}_f^{\text{b}}$, and a
\emph{distinguishable representation}, denoted by $\mathbf{v}_f^{\text{d}}$.
The basic representation $\mathbf{v}_f^{\text{b}}$ contains basic semantic information for matching a group of law articles that may apply to the case.
In contrast, the distinguishable representation $\mathbf{v}_f^{\text{d}}$ captures features that can effectively distinguish confusing law articles.
The concatenation of $\mathbf{v}_f^{\text{b}}$ and $\mathbf{v}_f^{\text{d}}$ is fed into subsequent classifiers to predict the labels of the JLP task.

As we mentioned, it is easy to distinguish dissimilar law articles as sufficient distinctions exist, and the difficulty in solving confusing charges lies in extracting distinguishable features of similar law articles.
To obtain the basic representation $\mathbf{v}_f^{\text{b}}$, therefore, we use one of the popular document encoding methods (e.g., CNN encoder \cite{kim2014convolutional} and Bi-RNN encoder \cite{yang2016hierarchical}).
To learn the distinguishable representation $\mathbf{v}_f^{\text{d}}$, we use a \emph{law distillation module} first to divide law articles to several communities to ensure that the law articles in each community are highly similar, and then extract each community $i$'s distinction vector (or, distinguishable features) $\mathbf{\beta}_i$ from the basic representation of law articles in community $i$.
Given the case's fact description,
from all communities' distinction vectors,
we select the most relevant one (i.e., $\mathbf{\beta}_{\hat{c}}$ in Fig.~\ref{fig:3}(a)) for attentively extracting the distinguishable features $\mathbf{v}_f^{\text{d}}$ in the \emph{fact re-encode module}.
In the follows, we elaborate law distillation module (Sec.~\ref{sec:law_interaction}) and fact re-encode module (Sec.~\ref{sec:re_encoder}) respectively.

\subsection{Distilling Law Articles} \label{sec:law_interaction}
A case might be misjudged due to the high similarity of some law articles.
To alleviate this problem, we design a law distillation module (cf. Fig.~\ref{fig:3} b) to extract distinguishable and representative information from all law  articles.
Specifically, it first uses a \emph{graph construction layer} (\textbf{GCL}) to divide law articles into different communities.
For each law article community,
a graph distillation layer is applied to learn its discriminative representation, hereinafter, called \textit{distinction vector}.

\subsubsection{Graph Construction Layer}
To find probably confusing law articles, we first construct a fully-connected graph $G^*$ for all law articles $\mathcal{L}$,
where the weight on the edge between a pair of law article $L_i, L_j\in \mathcal{L}$ is defined as the cosine similarity between their TF-IDF (Term Frequency-Inverse Document Frequency) representations $\mathbf{tf\_idf}_i$ and $\mathbf{tf\_idf}_j$.
Since confusing law articles are usually semantically similar
and there exists sufficient information to distinguish dissimilar law articles, we remove the edges with weights less than a predefined threshold $\tau$ from graph $G^*$.
By setting an appropriate $\tau$, we obtain a new graph $G = \{g_i\}_{i = 1}^{M}$ composed of several disconnected subgraphs $g_1, \ldots, g_M$ (or, communities),
where each $g_i, i=1,\ldots, M$ contains a specific community of probably confusing articles.
Our later experimental results demonstrate that this easy-to-implement method
effectively improves the performance of LADAN.

\subsubsection{Graph Distillation Layer}
To extract the distinguishable information from each community $g_i$, a straightforward way is to delete duplicate words and sentences presented in law articles within the community (as described in Sec.~\ref{sec:introduction}).
In addition to introducing significant errors, this simple method cannot be plugged into end-to-end neural architectures due to its non-differentiability.
To overcome the above issues, inspired by the popular graph convolution operator (\textbf{GCO})~\cite{kipf2016semi,hamilton2017inductive,velivckovic2017graph}, we propose a \emph{graph distillation operator} (\textbf{GDO}) to effectively extract distinguishable features.
Different from GCO, which computes the message propagation between neighbors and aggregate these messages to enrich representations of nodes in the graph, the basic idea behind our GDO is to learn effective features with distinction by removing similar features between nodes.

Specifically, for an arbitrary law article $L_i$, GDO uses a trainable weight matrix $\Psi$ to capture similar information between it and its neighbors in graph $G$, and a matrix $\Phi$ to extract effective semantic features of $L_i$.
At each layer $l\ge 0$, the aggregation of similar information between $L_i$ and its neighbors is removed from its representation, that is,
$$
 \mathbf{v}_{L_i}^{(l+1)} = \Phi^{(l)}\mathbf{v}_{L_i}^{(l)} - \sum_{L_j\in N_i}\frac{\Psi^{(l)}[\mathbf{v}_{L_i}^{(l)}, \mathbf{v}_{L_j}^{(l)}]}{|N_i|} + \mathbf{b}^{(l)}
$$
where $\mathbf{v}_{L_i}^{(l)} \in \mathbb{R}^{d_l}$ refers to the representation of law $L_i$ in the $l^\text{th}$ graph distillation layer, $N_i$ refers to the neighbor set of $L_i$ in graph $G$, $\mathbf{b}^{(l)}$ is the bias, and $\Phi^{(l)} \in \mathbb{R}^{d_{l+1} \times d_{l}}$ and $\Psi^{(l)} \in \mathbb{R}^{d_{l+1} \times 2d_{l}}$ are the trainable self weighted matrix and the neighbor similarity extracting matrix respectively. Note that $d_l$ is the dimension of the feature vector in the $l^\text{th}$ graph distillation layer.
We set $d_0 = d_s$, where ${d_s}$ is the dimension of basic representations $\mathbf{v}_f^{\text{b}}$ and $\mathbf{v}_{L_i}$. Similar to GCO, our GDO also supports multi-layer stacking.

Using GDO with $H$ layers, we output law article representation of the last layer, i.e., $\mathbf{v}_{L_i}^{(H)} \in \mathbb{R}^{d_H}$, which contains rich distinguishable features that can distinguish law article $L_i$ from the articles within the same community.
To further improve law articles' distinguishable features,
for each subgraph $g_i, i = 1,2, \dots, M$ in graph $G$,
we compute its distinction vector $\beta_i$
by using pooling operators to aggregate the distinguishable features of articles in $g_i$. Formally, $\beta_i$ is computed as:
$$
 \beta_i = [\text{MaP}(\{\mathbf{v}_{L_i}^{(H)}\}_{L_j\in g_i}), \text{MiP}(\{\mathbf{v}_{L_i}^{(H)}\}_{L_j\in g_i})]
$$
where $\text{MaP}(\cdot)$ and $\text{MiP}(\cdot)$ are the element-wise max pooling and element-wise min pooling operators respectively.

\subsection{Re-encoding Fact with Distinguishable Attention}
\label{sec:re_encoder}
To capture a law case's distinguishable features from its fact description $f$, we firstly define the following linear function,
which is used to predict its most related community $g_{\hat{c}}$ in graph $G$:
\begin{equation}\label{eq:graph_selection}
\hat{\mathbf{X}} = \text{softmax}(\mathbf{W}_g\mathbf{v}_f^{\text{b}} + \mathbf{b}_g)
\end{equation}
where $\mathbf{v}_f^{\text{b}}$ is the basic representation of fact description $f$,
$\mathbf{W}_g \in \mathbb{R}^{M \times d_s}$ and $\mathbf{b}_g \in \mathbb{R}^M$ are the trainable weight matrix and bias respectively. Each element $\hat{X}_i \in \hat{\mathbf{X}}$, $i=1,...,M$ reflects the closeness between fact description $f$ and law articles community $g_i$.
The most relevant community $g_{\hat{c}}$ is computed as
$$\hat{c} =\arg\max_{i=1,\ldots, M} \hat{X}_i.$$
Then, we use the corresponding community's distinction vector $\beta_{\hat{c}}$ to attentively extract distinguishable features from fact description $f$.

Inspired by~\cite{yang2016hierarchical}, we attentively extract distinguishable features based on word-level and sentence-level Bi-directional Gated Recurrent Units (Bi-GRUs).
Specifically, for each input sentence $S_i = [w_{i,1}, \cdots, w_{i, n_{i}}]$ in fact description $f$,  word-level Bi-GRUs will output a hidden state sequence, that is,
$$ \mathbf{h}_{i,j} = [\overrightarrow{\text{GRU}}(\mathbf{w}_{i,j}),  \overleftarrow{\text{GRU}}(\mathbf{w}_{i,j})],\ \ j = 1,..., n_i,$$
where $\mathbf{w}_{i,j}$ represents the word embedding of word $w_{i.j}$ and
$\mathbf{h}_{i,j} \in \mathbb{R}^{d_w}$.
Based on this hidden state sequence and the distinction vector $\beta_{\hat{c}}$, we calculate an attentive vector $[\alpha_{i,1},\ldots,\alpha_{i,n_i}]$,
where each $\alpha_{i,j}$ evaluates the discrimination ability of word $w_{i,j} \in S_i$. $\alpha_{i,j}$ is formally computed as:
$$\alpha_{i,j} =\frac {\exp(\text{tanh}(\mathbf{W}_w{\mathbf{h}}_{i,j})^\mathsf{T}(\mathbf{W}_{gw}\beta_{\hat{c}}))} {\sum_{j}\exp(\text{tanh}(\mathbf{W}_w{\mathbf{h}}_{i,j}) ^\mathsf{T}(\mathbf{W}_{gw}\beta_{\hat{c}}))},$$
where $\mathbf{W}_{w}$ and $\mathbf{W}_{gw}$ are trainable weight matrices.
Then, we get a representation of sentence $S_i$ as:
$$\mathbf{v}_{s_i} = \sum_{j = 1}^{n_i}\alpha_{i,j}\mathbf{h}_{i,j},$$
where $n_i$ denotes the word number in sentence $S_i$.

By the above word-level Bi-GRUs, we get a sentence representations sequence $[\mathbf{v}_{s_1}, \ldots, \mathbf{v}_{s_{n_f}}]$,
where $n_f$ refers to the number of sentences in the fact description $f$.
Based on this sequence, similarly, we build sentence-level Bi-GRUs and calculate a sentence-level attentive vector $[\alpha_{1},\ldots,\alpha_{n_f}]$ that reflects the discrimination ability of each sentence, and then get the fact's distinguishable representation $\mathbf{v}_f^{\text{d}} \in \mathbb{R}^{d_s}$.
Our sentence-level Bi-GRUs are formulated as:
$$ \mathbf{h}_{i} = [\overrightarrow{\text{GRU}}(\mathbf{v}_{s_i}),  \overleftarrow{\text{GRU}}(\mathbf{v}_{s_i})],\ \ i = 1,2,..., n_f, $$
$$\alpha_{i} =\frac {\exp(\text{tanh}(\mathbf{W_s}{\mathbf{h}}_{i})^\mathsf{T}(\mathbf{W}_{gs}\beta_{\hat{c}}))} {\sum_{i}\exp(\text{tanh}(\mathbf{W_s}{\mathbf{h}}_{i})^\mathsf{T}(\mathbf{W}_{gs}\beta_{\hat{c}}))},$$
$$
\mathbf{v}_f^{\text{d}} = \sum_{i}\alpha_{i}\mathbf{h}_{i}.
$$

\subsection{Prediction and Training}\label{sec:training}
We concatenate the basic representation $\mathbf{v}_f^{\text{b}}$ and the distinguishable representation $\mathbf{v}_f^{\text{d}}$ as the final representation of fact description $f$, i.e., $\tilde{\mathbf{v}}_f = [\mathbf{v}_f^{\text{b}}, \mathbf{v}_f^{\text{d}}]$.
Based on $\tilde{\mathbf{v}}_f$,
we generate a corresponding feature vector $\tilde{\mathbf{v}}_f^j$ for each subtask $t_j$, $j=1,2,3$ mentioned in Sec.~\ref{sec:problem}, i.e., $t_1$: \textbf{law article prediction}; $t_2$: \textbf{charge prediction}; $t_3$: \textbf{term of penalty prediction}.
To obtain the prediction for each subtask,
we use a linear classifier:
$$ \hat{y}_{j} = \text{softmax}(\mathbf{W}_p^j \tilde{\mathbf{v}}_f^j + \mathbf{b}_p^j), $$
where $\mathbf{W}_p^j$ and $\mathbf{b}_p^j$ are parameters specific to task $t_j$.
For training, we compute a cross-entropy loss function for each subtask and take the loss sum of all subtasks as the overall prediction loss:
$$ \mathscr{L}_p = -\sum_{j=1}^3 \sum_{k=1}^{\lvert Y_j\lvert} {y}_{j,k} \log(\hat{y}_{j,k}),$$
where $\lvert Y_j\lvert $ denotes the number of different classes (or, labels) for task $t_j$ and $[y_{j,1}, y_{j,2}, \dots, y_{j,\lvert Y_j\lvert}]$  refers to the ground-truth vector of task $t_j$.
Besides, we also consider the loss of law article community prediction (i.e., Eq.~\ref{eq:graph_selection}):
$$ \mathscr{L}_{c} = -\lambda \sum_{j=1}^{M} {X}_{j}\log({\hat{X}_{j}}),$$
where $[X_1, X_2,\dots, X_M]$ is the ground-truth vector of the community including the correct law article applied to the law case.
In summary, our final overall loss function is:
\begin{equation}\label{eq:loss}
 \mathscr{Loss} = \mathscr{L}_p + \mathscr{L}_{c}
\end{equation}

\section{Experiments} \label{sec:experiment}
\subsection{Datasets}
To evaluate the performance of our method,
we use the publicly available datasets of
the \textbf{C}hinese \textbf{AI} and \textbf{L}aw challenge (CAIL2018)\footnote{\url{http://cail.cipsc.org.cn/index.html}}~\cite{xiao2018cail2018}: \emph{CAIL-small} (the exercise stage dataset) and \emph{CAIL-big}  (the first stage dataset).
The case samples in both datasets contain fact description, applicable law articles, charges, and the terms of penalty.
For data processing, we first filter out samples with fewer than 10 meaningful words.
To be consistent with state-of-the-art methods, we filter out the case samples with multiple applicable law articles and multiple charges.
Meanwhile, referring to~\cite{zhong2018legal}, we only keep the law articles and charges that apply to not less than 100 corresponding case samples and divide the terms of penalty into non-overlapping intervals.
The detailed statistics of the datasets are shown in Table~\ref{tab:data}.

\setlength{\abovecaptionskip}{0.1cm}
\setlength{\belowcaptionskip}{-0.2cm}
\begin{table}[htp]
\small
\centering
\begin{tabular}{@{}lrr@{}}
\toprule
Dataset & CAIL-small & CAIL-big \\
\midrule
\#Training Set Cases  & 101,619    & 1,587,979   \\
\#Test Set Cases  & 26,749    & 185,120  \\
\#Law Articles  & 103   & 118  \\
\#Charges     & 119     & 130  \\
\#Term of Penalty     & 11    & 11  \\
\bottomrule
\end{tabular}
\caption{Statistics of datasets.}\label{tab:data}
\end{table}

\subsection{Baselines and Settings}
\paragraph{Baselines.} We compare LADAN with some baselines, including:
\begin{itemize}
\item \textbf{CNN~\cite{kim2014convolutional}:} a CNN-based model with multiple filter window widths for text classification.
\item \textbf{HARNN~\cite{yang2016hierarchical}:} an RNN-based neural network with a hierarchical attention mechanism for document classification.
\item \textbf{FLA~\cite{luo2017learning}:} a charge prediction method that uses an attention mechanism to capture the interaction between fact description and applicable laws.
\item \textbf{Few-Shot~\cite{hu2018few}:} a discriminating confusing charge method, which extracts features about ten predefined attributes from fact descriptions to enforce semantic information.

\item \textbf{TOPJUDGE~\cite{zhong2018legal}:} a topological multi-task learning framework for LJP, which formalizes the explicit dependencies over subtasks in a directed acyclic graph.
\item \textbf{MPBFN-WCA~\cite{yang2019legal}:} a multi-task learning framework for LJP with multi-perspective forward prediction and backward verification, which is the state-of-the-art method.
\end{itemize}

Similar to existing works~\cite{luo2017learning,zhong2018legal}, we train the baselines CNN, HLSTM and FLA using a multi-task framework (recorded as MTL) and select a set of the best experimental parameters according to the range of the parameters given in their original papers.
Besides, we use our method LADAN with the same multi-task framework (i.e., Landan+MTL, LADAN+TOPJUDGE, and LADAN+MPBFN) to demonstrate our superiority in feature extraction.

\paragraph{Experimental Settings.} We use the THULAC~\cite{Thulac2016} tool to get the word segmentation because all case samples are in Chinese.
Afterward, we use the Skip-Gram model~\cite{mikolov2013distributed} to pre-train word embeddings on these case documents, where the model's embedding size and frequency threshold are set to 200 and 25 respectively.
Meanwhile, we set the maximum document length as 512 words for CNN-based models in baselines and set the maximum sentence length to 100 words and maximum document length to 15 sentences for LSTM-based models.
As for hyperparameters setting, we set the dimension of all latent states (i.e., $d_w$, $d_s$, $d_l$ and $d_f$) as 256 and the threshold $\tau$ as $0.3$.
In our method LADAN, we use two graph distillation layers, and a Bi-GRU with a randomly initialized attention vector $u$ is adopted as the basic document encoder.
For training, we set the learning rate of Adam optimizer to $10^{-3}$, and the batch size to 128.
After training every model for 16 epochs, we choose the best model on the validation set for testing.\footnote{Our source codes are available at \url{https://github.com/prometheusXN/LADAN}}

\begin{table*}[t]
\centering
\resizebox{\textwidth}{!}{%
\begin{tabular}{lcccccccccccc}
\toprule
 Tasks              & \multicolumn{4}{c}{Law Articles}                     & \multicolumn{4}{c}{Charges}                          & \multicolumn{4}{c}{Term of Penalty}                   \\ \cmidrule(lr){2-5} \cmidrule(lr){6-9} \cmidrule(lr){10-13}
 Metrics            & Acc.        & MP          & MR          & F1          & Acc.        & MP          & MR          & F1          & Acc.        & MP          & MR          & F1          \\
 \midrule
 FLA+MTL            & $77.74$     & $75.32$     & $74.36$     & $72.93$     & $80.90$     & $79.25$     & $77.61$     & $76.94$     & $36.48$     & $30.94$     & $28.40$     & $28.00$     \\
 CNN+MTL            & $78.71$     & $76.02$     & $74.87$     & $73.79$     & $82.41$     & $81.51$     & $79.34$     & $79.61$     & $35.40$     & $33.07$     & $29.26$     & $29.86$     \\
 HARNN+MTL          & $79.79 $    & $75.26$     & $76.79$     & $74.90$     & $83.80$     & $82.44$     & $82.78$     & $82.12$     & $36.17$     & $34.66$     & $31.26$     & $31.40$     \\
 Few-Shot+MTL       & $79.30$     & $77.80$     & $77.59$     & $76.09$     & $83.65$     & $80.84$     & $82.01$     & $81.55$     & $36.52$     & $35.07$     & $26.88$     & $27.14$     \\
 TOPJUDGE           & $79.88$     & $79.77$     & $73.67$     & $73.60$     & $82.10$     & $83.60$     & $78.42$     & $79.05$     & $36.29$     & $34.73$     & $32.73$     & $29.43$     \\
 MPBFN-WCA          & $79.12$     & $76.30$     & $76.02$     & $74.78$     & $ 82.14$    & $82.28$     & $80.72$     & $80.72$     & $36.02$     & $31.94$     & $28.60$     & $29.85$     \\
 \midrule
 \bf{LADAN+MTL}      & $\bf 81.20$ & $\bf 78.24$ & $\bf 77.38$ & $\bf 76.47$ & $\bf 85.07$ & $\bf 83.42$ & $\bf 82.52$ & $\bf 82.74$ & $\bf 38.29$ & $\bf 36.16$ & $\bf 32.49$ & $\bf 32.65$ \\
 \bf{LADAN+TOPJUDGE} & $\bf 81.53$ & $\bf 78.62$ & $\bf 78.29$ & $\bf 77.10$ & $\bf 85.12$ & $\bf 83.64$ & $\bf 83.57$ & $\bf 83.14$ & $\bf 38.34$ & $\bf 36.39$ & $\bf 32.75$ & $\bf 33.53$ \\
 \bf{LADAN+MPBFN}    & $\bf 82.34$ & $\bf 78.79$ & $\bf 77.59$ & $\bf 76.80$ & $\bf 84.83 $ & $\bf 83.33$ & $\bf 82.80$ & $\bf 82.85$ & $\bf 39.35$ & $\bf 36.94$ & $\bf 33.25$ & $\bf 34.05$ \\
 \midrule
 \bottomrule
\end{tabular}%
}
\caption{Judgment prediction results on CAIL-small.}\label{tab:CAIL-small}
\end{table*}

\begin{table*}[t]
\centering
\resizebox{\textwidth}{!}{%
\begin{tabular}{lcccccccccccc}
\toprule
 Tasks              & \multicolumn{4}{c}{Law Articles}                     & \multicolumn{4}{c}{Charges}                          & \multicolumn{4}{c}{Term of Penalty}                   \\ \cmidrule(lr){2-5} \cmidrule(lr){6-9} \cmidrule(lr){10-13}
 Metrics            & Acc.        & MP          & MR          & F1          & Acc.        & MP          & MR          & F1          & Acc.        & MP          & MR          & F1          \\
 \midrule
 FLA+MTL   &   $93.23 $ & $ 72.78 $ & $64.30 $ & $66.56$ & $92.76$ & $76.35$ & $68.48$ & $70.74$ & $57.63$ & $48.93$ & $45.00$ & $46.54$ \\
 CNN+MTL   & $95.84$ & $83.20$ & $75.31$ & $77.47$ & $95.74$ & $86.49$ & $79.00$ & $81.37$ & $55.43$ & $45.13$ & $38.85$ & $39.89$
 \\
 HARNN+MTL &  $95.63$ & $81.48$ & $74.57$ & $77.13$ & $95.58$ & $85.59$ & $79.55$ & $81.88$ & $57.38$ & $43.50$ & $40.79$ & $42.00$
 \\
 Few-Shot+MTL  & $96.12$    & $85.43$   & $80.07$   & $81.49$   & $96.04$   & $88.30$   & $80.46$   & $83.88$   & $57.84$   & $47.27$   & $42.55$   & $43.44$ \\
 TOPJUDGE &  $95.85$ & $84.84$ & $74.53$ & $77.50$ & $95.78$ & $86.46$ & $78.51$ & $81.33$ & $57.34$ & $47.32$ & $42.77$ & $44.05$
 \\
 MPBFN-WCA & $96.06$ & $85.25$ & $74.82$ & $78.36$ & $95.98$ & $89.16$ & $79.73$ & $83.20$ & $58.14$ & $45.86$ & $39.07$ & $41.39$ \\
 \midrule
 \bf LADAN+MTL & $\bf 96.57$ & $\bf 86.22$ & $\bf 80.78$ & $\bf 82.36$ & $\bf 96.45$ & $\bf 88.51$ & $\bf 83.73$ & $\bf 85.35$ & $\bf 59.66$ & $\bf 51.78$ & $\bf 45.34$ & $\bf 46.93$ \\
 \bf LADAN+TOPJUDGE & $\bf 96.62$ & $\bf 86.53$ & $\bf 79.08$ & $\bf 81.54$ & $\bf 96.39$ & $\bf 88.49$ & $\bf 82.28$ & $\bf 84.64$ & $\bf 59.70$ & $\bf 51.06$ & $\bf 45.46$ & $\bf 46.96$ \\
\bf LADAN+MPBFN &  $\bf 96.60$ & $\bf 86.42$ & $\bf 80.37$ & $\bf 81.98$ & $\bf 96.42$ & $\bf 88.45$ & $\bf 83.08$ & $\bf 84.95$ & $\bf 59.85$ & $\bf 51.75$ & $\bf 45.59$ & $\bf 47.18$ \\
\midrule
 \bottomrule
\end{tabular}%
}
\caption{Judgment prediction results on CAIL-big.}\label{tab:CAIL-big}
\end{table*}

\subsection{Experimental Results}
To compare the performance of the baselines and our methods,
we choose four metrics that are widely used for multi-classification tasks, including accuracy (Acc.), macro-precision (MP), macro-recall (MR), and macro-F1 (F1).
Since the problem of confusing charges often occurs between a few categories, the main metric is the F1 score.
Tables~\ref{tab:CAIL-small} and~\ref{tab:CAIL-big} show the experimental results on datasets CAIL-small and CAIL-big, respectively. Our method LADAN performs the best in terms of all evaluation metrics.
Because both CAIL-small and CAIL-big are imbalanced datasets, we focus on comparing the F1-score, which more objectively reflects the effectiveness of our LADAN and other baselines.
Compared with the state-of-the-art MPBFN-WCA,
LADAN improved the F1-scores of law article prediction, charge prediction, and term of penalty prediction on dataset CAIL-small by $2.02$\%, $2.42$\% and $4.20$\% respectively,
and about $3.18$\%, $1.44$\% and $5.79$\% on dataset CAIL-big.
Meanwhile, the comparison under the same multi-task framework (i.e., MTL, TOPJUDGE, and MPBFN) shows that our LADAN extracted more effective features from fact descriptions than all baselines.
Meanwhile, we can observe that the performance of Few-shot on charge prediction is close to LADAN, but its performance on the term of penalty prediction is far from ideal. It is because the ten predefined attributes of Few-Shot are only effective for identifying charges, which also proves the robustness of our LADAN.
The highest MP- and MR-scores of LADAN also demonstrates its ability to distinguish confusing law articles.
Note that all methods' performance on dataset CAIL-big is better than that on CAIL-small, which is because the training set on CAIL-big is more adequate.

\subsection{Ablation Experiments}
To further illustrate the significance of considering the difference between law articles, we conducted ablation experiments on model \textit{LADAN+MTL}  with dataset CAIL-small.
To prove the effectiveness of our graph construction layer (\textbf{GCL}), we build a LADAN model with the GCL's removing threshold $\tau=0$ (i.e., ``-no GCL'' in Table ~\ref{tab:ablation_exp}), which directly applies the GDO on the fully-connected graph $G^*$ to generate a global distinction vector $\beta_g$ for re-encoding the fact description.
To verify the effectiveness of our graph distillation operator (\textbf{GDO}), we build a no-GDO LADAN model (i.e., ``-no GDO'' in Table ~\ref{tab:ablation_exp}), which directly pools each subgraph $g_i$ to a distinction vector $\beta_i$ without GDOs.
To evaluate the importance of considering the difference among law articles, we remove both GCL and GDO from LADAN by setting $\tau=1.0$ (i.e., ``-no both'' in Table ~\ref{tab:ablation_exp}), i.e., each law article independently extracts the attentive feature from fact description.
In Table~\ref{tab:ablation_exp},
we see that both GCL and GDO effectively improve the performance of LADAN.
GCL is more critical than GDO because GDO has a limited performance when the law article communities obtained by GCL are not accurate.
When removing both GCL and GDO, the accuracy of LADAN decreases to that of HARNN+MTL, which powerfully demonstrates the effectiveness of our method exploiting differences among similar law articles.

\begin{table}[t]
\centering
\resizebox{\linewidth}{!}{%
\begin{tabular}{@{}lrrrrrr@{}}
\toprule
Tasks & \multicolumn{2}{c}{Law} & \multicolumn{2}{c}{Charge} & \multicolumn{2}{c}{Penalty} \\ \cmidrule(lr){2-3} \cmidrule(lr){4-5} \cmidrule(lr){6-7}
Metrics & \multicolumn{1}{c}{Acc.} & \multicolumn{1}{c}{F1} & \multicolumn{1}{c}{Acc.} & \multicolumn{1}{c}{F1} & \multicolumn{1}{c}{Acc.} & \multicolumn{1}{c}{F1} \\ \midrule
LADAN+MTL & $ 81.20 $ & $ 76.47 $ & $ 85.07 $ & $ 83.14 $ & $ 38.29 $ & $ 32.65 $ \\
-no GCL & $ 80.46 $ & $ 75.98 $ & $ 84.04 $ & $ 82.33 $ & $ 37.80 $ & $ 31.85 $ \\
-no GDO & $ 80.82 $ & $ 76.19 $ & $ 84.65 $ & $ 82.50 $ & $ 36.69 $ & $ 31.62 $ \\
-no both & $ 79.79 $ & $ 74.97 $ & $ 83.72 $ & $ 82.02 $ & $ 34.87 $ & $ 31.34 $ \\ \bottomrule
\end{tabular}%
}
\caption{Ablation analysis on CAIL-small.}\label{tab:ablation_exp}
\end{table}

\begin{figure*}[t]
\centering
\includegraphics[width=0.98\linewidth]{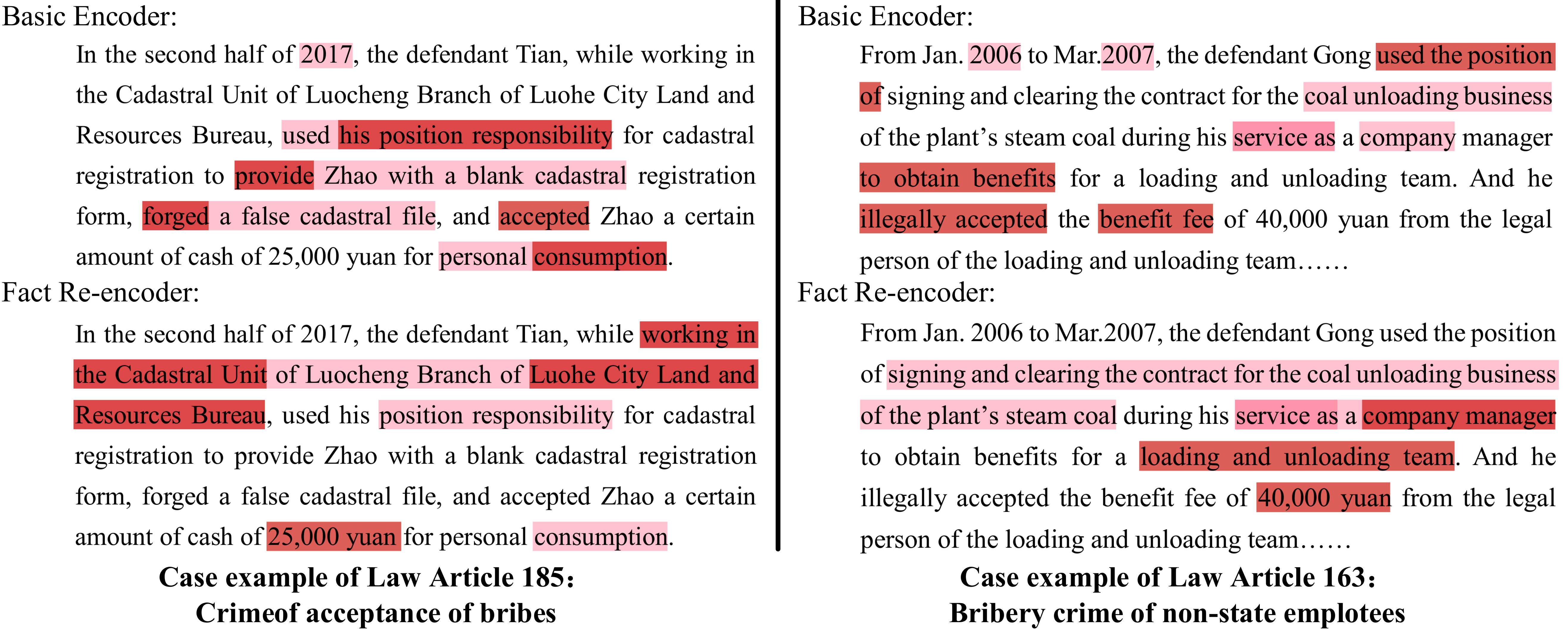}
\caption{The attention visualization on case examples for Article 185 and Article 163.}
\label{fig:case_study}
\end{figure*}

\subsection{Case Study} \label{sec:case_study}
To intuitively verify that LADAN effectively extracts distinguishable features, we visualize the attention of LADAN's encoders.
Figure~\ref{fig:case_study} shows two law case examples, each for \textit{Article 385} and \textit{Article 163}, respectively, where the darker the word is, the higher the attention weight it gets in the corresponding encoder, i.e., its information is more important to the encoder.
For the basic encoder, we see that the vital information in these two cases is very similar, which both contain the word like \textit{``use position''} \textit{``accept benefit''} \textit{``accept ... cash''}, etc. Therefore, when using just the representation of basic encoder to predict acceptable law articles, charges and terms of penalty, these two cases tend to be misjudged.
As we mentioned in Sec.~\ref{sec:re_encoder}, with the distinction vector, our fact re-encoder focuses on extracting distinguishable features like defendants' identity information (e.g., \textit{``company manager'' ``working in the Cadastral Unit of Luocheng Branch of Luohe City Land and Resources Bureau''} in our examples), which effectively distinguish the applicable law articles and charges of these two cases.

\section{Conclusion} \label{sec: conclusion}
In this paper, we present an end-to-end model, LADAN, to solve the issue of confusing charges in LJP.
In LADAN, a novel attention mechanism is proposed to extract the key features for distinguishing confusing law articles attentively.
Our attention mechanism not only considers the interaction between fact description and law articles but also the differences among similar law articles, which are effectively extracted by a graph neural network GDL proposed in this paper.
The experimental results on real-world datasets show that our LADAN raises the F1-score of state-of-the-art by up to $5.79$\%.
In the future, we plan to study complicated situations such as a law case with multiple defendants and charges.

\section*{Acknowledgments}

The research presented in this paper is supported in part by National Key R\&D Program of China (2018YFC0830500),Shenzhen Basic Research Grant (JCYJ20170816100819428), National Natural Science Foundation of China (61922067, U1736205, 61902305), MoE-CMCC “Artifical Intelligence” Project (MCM20190701), National Science Basic Research Plan in Shaanxi Province of China (2019JM-159), National Science Basic Research Plan in Zhejiang Province of China (LGG18F020016).

\bibliography{anthology,acl2020}
\bibliographystyle{acl_natbib}

\end{document}